\documentclass{article}
\usepackage{spconf,amsmath,graphicx}
\usepackage[hyphens]{url}

\title{Uncovering communities of pipelines in the task-fMRI analytical space}

\name{Elodie Germani$^{1}$ \qquad Elisa Fromont$^{2\dagger}$ \qquad Camille Maumet$^{1\dagger}$}
  
\address{$^{1}$ Univ Rennes, Inria, CNRS, Inserm, Rennes, France \\
$^{2}$Univ Rennes, IUF, Inria, CNRS, Rennes, France \\ 
\ninept $^{\dagger}$ Joint senior authorship}

\ninept

\begin{document}

\maketitle
\begin{abstract}
Analytical workflows in functional magnetic resonance imaging are highly flexible with limited best practices as to how to choose a pipeline. While it has been shown that the use of different pipelines might lead to different results, there is still a lack of understanding of the factors that drive these differences and of the stability of these differences across contexts. We use community detection algorithms to explore the pipeline space and assess the stability of pipeline relationships across different contexts. We show that there are subsets of pipelines that give similar results, especially those sharing specific parameters (e.g. number of motion regressors, software packages, etc.). Those pipeline-to-pipeline patterns are stable across groups of participants but not across different tasks. By visualizing the differences between communities, we show that the pipeline space is mainly driven by the size of the activation area in the brain and the scale of statistic values in statistic maps. 

\end{abstract}
\begin{keywords}
neuroimaging, pipeline, variability, communities, stability
\end{keywords}
\section{Introduction}
\label{sec:intro}

Functional Magnetic Resonance Imaging (fMRI) is a neuroimaging technique that measures brain activity associated with a specific task (activity performed by the participant during the acquisition) or cognitive paradigm (mental processes involved during the task). A typical fMRI data analysis can be split into three main steps: pre-processing, subject-level (first-level), and group-level (second-level) statistics. The sequence of steps performed during the analysis is referred to as a pipeline. Results from these pipelines are 3-dimensional statistic maps representing the activation of the brain for a contrast of interest, which focus on a particular aspect of the task, at the subject-level (individual activation patterns) or at the group-level (activation patterns across multiple subjects). These statistic maps can be used for instance to perform brain decoding~\cite{firat_deep_2014} (\textit{i.e.} identifying stimuli and cognitive states from brain activities) or studying brain disorders~\cite{yin_deep_2022} (\textit{i.e.} identifying differences in activation patterns caused by brain disorders).

A large number of software packages and methods can be used at each step, making the choice of fMRI pipeline a challenging process for practitioners.
As an illustration, in a meta-analysis~\cite{carp_secret_2012}, Carp examined the different pipelines used in $240$ fMRI studies and found over $200$ unique pipelines. Recent studies \cite{botvinik2020variability, bowring_maumet} demonstrated the impact of using different pipelines in the results of an fMRI study. For instance, in \cite{botvinik2020variability}, 70 research teams analyzed the same fMRI dataset using their favorite pipeline. Overall the study found substantial differences in statistic maps and conclusions to binary research hypotheses across teams (e.g. about the activation of the brain in a particular area during a particular task). Yet, there is still little understanding as to which factor in the fMRI pipelines are the main drivers of that variability. 

In an effort to guide practitioners into the pipeline space, Dafflon et al.~\cite{dafflon2022guided} proposed a new method to identify subsets of pipelines that are best suited to answer a problem for which ground truths are available, such as predicting the age of participants. 

In practice, researchers commonly explore multiple valid analytic alternatives, but often report their results relative only to a single pipeline (or to a few set of variants). This selective reporting can result in an increase of false positive findings~\cite{ioannidis_why_2008, simmons_false-positive_2011,gelman_garden_nodate}. Multiverse analyses~\cite{steegen_increasing_2016}, which report the results of an experiment carried out under different analytic conditions, are a promising solution to improve the reliability of neuroimaging studies and to limit selective reporting. But a systematic investigation of the pipeline space is impractical due to the high number of possible pipelines. To improve our knowledge of the pipeline space, it is necessary to find a way to measure distances between different analysis methods. Such relationship measurements can facilitate the selection of a set of pipeline parameters that are the main drivers of variability in the result space. The definition of this pre-defined set of pipelines to test would improve the quality of the results of the multiverse analysis, but also decrease the computational time required for such experiment.

Investigating the pipeline space can also help in understanding the homogeneity (i.e., pipelines that give similar results) but also the heterogeneity (i.e. pipelines that have a different behavior) of the pipelines. Rolland et al.~\cite{rolland_towards_2022} recently showed the problems arising when combining subject-level results obtained from different pipelines for group-level analyses. As we may expect to see a growing proportion of derived data shared on public platforms (for instance NeuroVault~\cite{gorgolewski_neurovaultorg_2015}), with subject-level or group-level statistic maps in the future, it is thus necessary to find a way to combine such data. As we can suppose that such issue is exacerbated for pipelines presenting more distant results, their identification using dedicated measurements would be a first step to help reproducibility through data re-use. 

One open question is whether patterns observed across pipelines are stable across different contexts: group of subjects, cognitive paradigm, acquisition parameters, etc. In a recent work~\cite{germani:inserm-04531405}, we proposed a method to combine results from different pipelines by converting them between pipelines using style transfer. Style transfer frameworks aim at learning a mapping between two domains and at applying this mapping to data. If the mapping is different between contexts (\textit{e.g.} different cognitive tasks), a framework trained to transfer statistic maps of a particular paradigm would not be applicable to other statistic maps. Exploring the stability of the relationships between pipeline results is thus of particular importance to assess the potential of our solution, and beyond of any solution that aims at being generalizable across different set of participants or fMRI cognitive tasks.

To identify these patterns and measure distances between pipelines, clustering algorithms can be applied to statistic maps. However, because the data are high-dimensional and suffer from large number of sources of variability at different level (at the subject and group-level as brain activity patterns differ across subjects, at the acquisition level since fMRI scanners and protocols often vary between centers and studies, etc.), distance measures between statistic maps are often meaningless and unrelated dimensions might mask existing clusters~\cite{parsons_subspace_2004}. In such case, subspace clustering algorithms are typically used to find clusters in different subspaces within a dataset. 

Here, we used community detection algorithms (i.e. clustering on graphs) to explore the pipeline space and assess the stability of pipeline results across different groups of subjects and cognitive paradigm. Using a clustering in two steps, we first look for clusters of pipelines for each group of subjects and explore how these clusters are similar across different groups. We aim at identifying groups of pipelines that are stable across different contexts (i.e. different contrasts or group of participants). If two pipelines are located in the same community (i.e. the two pipelines present similar results) in different contexts, we can consider that their relationship is relatively stable. 

To our knowledge, this is the first time that the task-fMRI pipeline space is represented as a graph, which takes into account different groups of subjects. We then apply a graph clustering method to study the pipeline space and to explore how relationships between pipeline results change depending on the study context. In this study, we used a recently built constrained multi-pipeline dataset to derive general findings on the effect of specific parameters on the distance between pipelines results and to improve our knowledge about the pipeline space. 

In the following section, we present the multi-pipeline dataset and explain our processing steps for the sake of reproducibility. We then present our analytical tools in Section~\ref{sec:method}. Section~\ref{sec:results} shows our results that we discuss in section~\ref{sec:discussion}. 

\section{Material} \label{sec:material}

Subject-level data were obtained from the HCP multi-pipeline dataset~\cite{hcp_multi_pipeline}. This dataset contains subject-level data from 1,080 subjects of the HCP Young Adult S1200 release~\cite{hcp_overview} whose unprocessed functional and structural data for the motor task were analyzed using 24 pipelines. It also contains group-level statistic maps for 1,000 groups of 50 subjects, randomly sampled from the 1,080 subjects. 

The pipelines varied on the following set of parameters:
\begin{itemize}
\item Software package: SPM (Statistical Parametric Mapping, RRID: SCR\_007037)~\cite{penny2011statistical} or FSL (FMRIB Software Library, RRID: SCR\_002823)~\cite{jenkinson2012fsl}.
\item Smoothing kernel: Full-Width at Half-Maximum (FWHM) was equal to either 5mm or 8mm. 
\item Number of motion regressors included in the General Linear Model (GLM) for the first-level analysis: 0, 6 (3~rotations, 3~translations) or 24 (the 6~previous regressors + 6~derivatives and the 12~corresponding squares of regressors).
\item Presence (1) or absence (0) of the derivatives of the Hemodynamic Response Function (HRF) in the GLM for the first-level analysis. Only the temporal derivatives were added in FSL pipelines and both the temporal and dispersion derivatives were for SPM pipelines.
\end{itemize}
More details about the subject-level pipelines can be found in~\cite{hcp_multi_pipeline}. In this dataset, pipelines implemented in the different software packages were aligned by changing the software package default values. In total, those combinations provided a set of 24~different subject-level pipelines ( 2~software packages $\times$ 2 smoothing kernels $\times$ 3~numbers of motion regressors $\times$ 2~HRF ).

In the following, we refer to these parameters respectively as `software', `FWHM', `motion regressors' and `HRF derivatives'. For instance, pipelines built using the FSL software, smoothing with a kernel FWHM of 8mm, no motion regressors and no HRF derivatives will be denoted by `fsl, 8, 0, 0'. 

In the dataset, all pipelines were applied to each subject for the different motor contrasts studied (i.e. right-hand, right-foot, left-hand, left-foot and tongue). For second-level analysis, as SPM and FSL provide equivalent statistical methods, we arbitrarily chose SPM. 1,000 groups of 50 participants were randomly sampled among the 1,080 participants, leading to 1,000 statistic maps for each pipeline and contrast. Second-level analyses were performed using SPM with default parameters. We used the same second-level analysis method for all pipelines in order to focus on differences in the first-level analyses. 

\section{Methods}\label{sec:method}

To study the stability of the pipeline results and of the relationships between results across different contexts, we computed graphs of similarity between the statistic maps of different pipelines for each group and used the \emph{Louvain} community detection algorithm \cite{Blondel_2008} to partition each graph. \emph{Stability} was measured for each pair of pipelines as the number of groups (out of 1,000) for which the two pipelines were located in the same community. Graphs and communities were computed using NumPy (RRID:SCR\_008633) and Networkx (RRID:SCR\_016864). The code produced to run the experiments and to create the figures and tables of this paper is available in the Software Heritage public archive \cite{code_swh}. 

\begin{figure}[h]
    \centering
    \includegraphics[width=7.5cm]{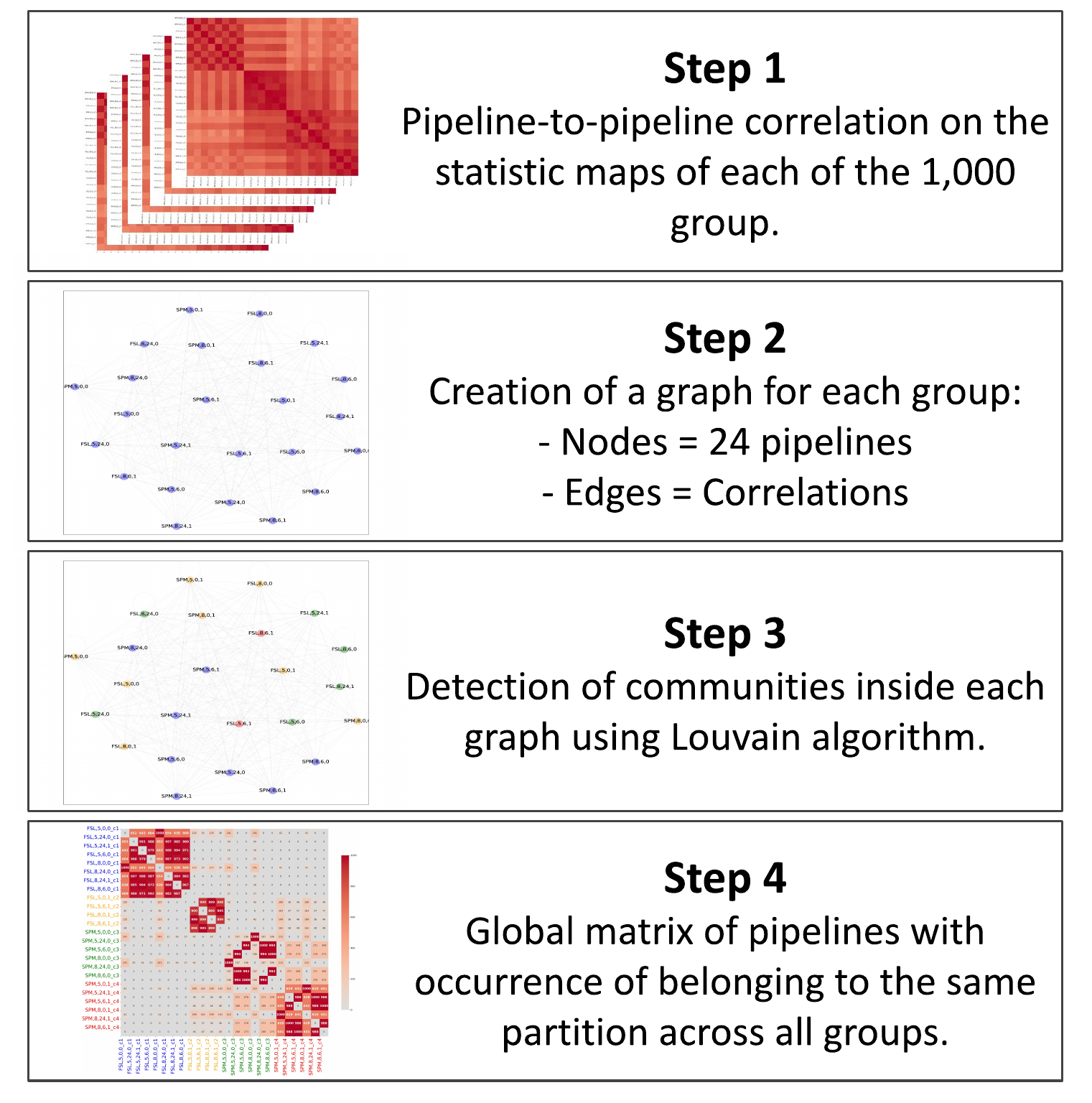}
    \caption{Workflow of community detection in the pipeline space across different groups of participants and contrasts.}
    \label{fig:fig1}
\end{figure}

\subsection{Data processing}

Group-level data obtained with different software packages did not have the same dimension, as standard MNI templates used for normalization are different between FSL and SPM. To be able to compute correlation between maps obtained with the two software packages, we had to resample group-level maps to a common dimension. We used Nilearn~\cite{abraham_machine_2014} (RRID: SCR\_001362) to resample all statistic maps from all pipelines to the MNI152Asym2009 brain template with a 2mm resolution using continuous interpolation. We computed a brain mask as the intersection of all group-level brain masks from all pipelines. This mask was also resampled to the MNI brain template using nearest-neighbors interpolation and applied to the resampled group-level data. In the end, group-level statistic maps from all pipelines were resampled to the same dimensions and masked using the same brain mask. 

\subsection{Graph computation and community detection}

We computed the similarity for each pair of pipelines in terms of Pearson's correlation coefficient between statistic maps (Fig.~\ref{fig:fig1} - Step 1) This correlation matrix was used as an adjacency matrix to build an undirected weighted multi-graph for each group, with nodes representing the statistic maps of the different pipelines (V = {`fsl,0,0,0', `fsl,0,0,1', etc.}) and edges labeled by the correlation coefficient between each pipeline (E={(`fsl,0,0,0',`fsl,0,0,1'), etc.}) (Fig.~\ref{fig:fig1} - Step 2). After computation, each graph was partitioned using the Louvain algorithm \cite{Blondel_2008} to detect the best partitions based on \emph{modularity} optimization (Fig.~\ref{fig:fig1} - Step 3), which represents the density of links inside communities as compared to links between communities. Therefore, the communities detected in each graph represent the pipelines that give similar results for the corresponding group. 

To explore the stability of the communities detected acrosss different groups of participants, we counted, for each pair of pipelines, the number of groups for which the two pipelines were located in the same community (Fig.~\ref{fig:fig1} - Step 4). A high value reflected a high similarity and stability across groups. This matrix was used to build a second graph, global across groups, in which nodes represent the different pipelines and edges represent the stability measure mentioned above. Louvain community detection algorithm was again applied to this second graph to detect communities in which pipeline provided similar statistic maps across different groups. These global graphs were computed for each contrast.

\subsection{Communities-specific features}

We visually assessed differences in the statistic maps in different communities by computing the mean statistic map of each pipeline across groups for each contrast. These were thresholded assuming a Standard Normal distribution (and effectively leading to conservative results) and  using a False Discovery Rate (FDR) of $p < 0.05$. For each pipeline map, we computed the number of activated voxels in the thresholded maps, but also within the Region of Interest (ROI) of the Primary Motor Cortex (M1), extracted from the probabilistic \emph{Jülich Atlas}, available in \emph{Nilearn} \cite{abraham_machine_2014} (RRID: SCR\_001362). This ROI is usually used to extract regional statistic values inside a whole-brain statistic brain of the motor task.
The goal was to identify the specific patterns of each community, to understand why a pipeline was located inside a community, and to explore the potential impact on the results of the pipelines. 

\section{Results} \label{sec:results}

\subsection{Characteristics of the pipeline space for contrast \textit{right-hand}}

\begin{figure}[htb]
    \centering
    \includegraphics[width=8.5cm]{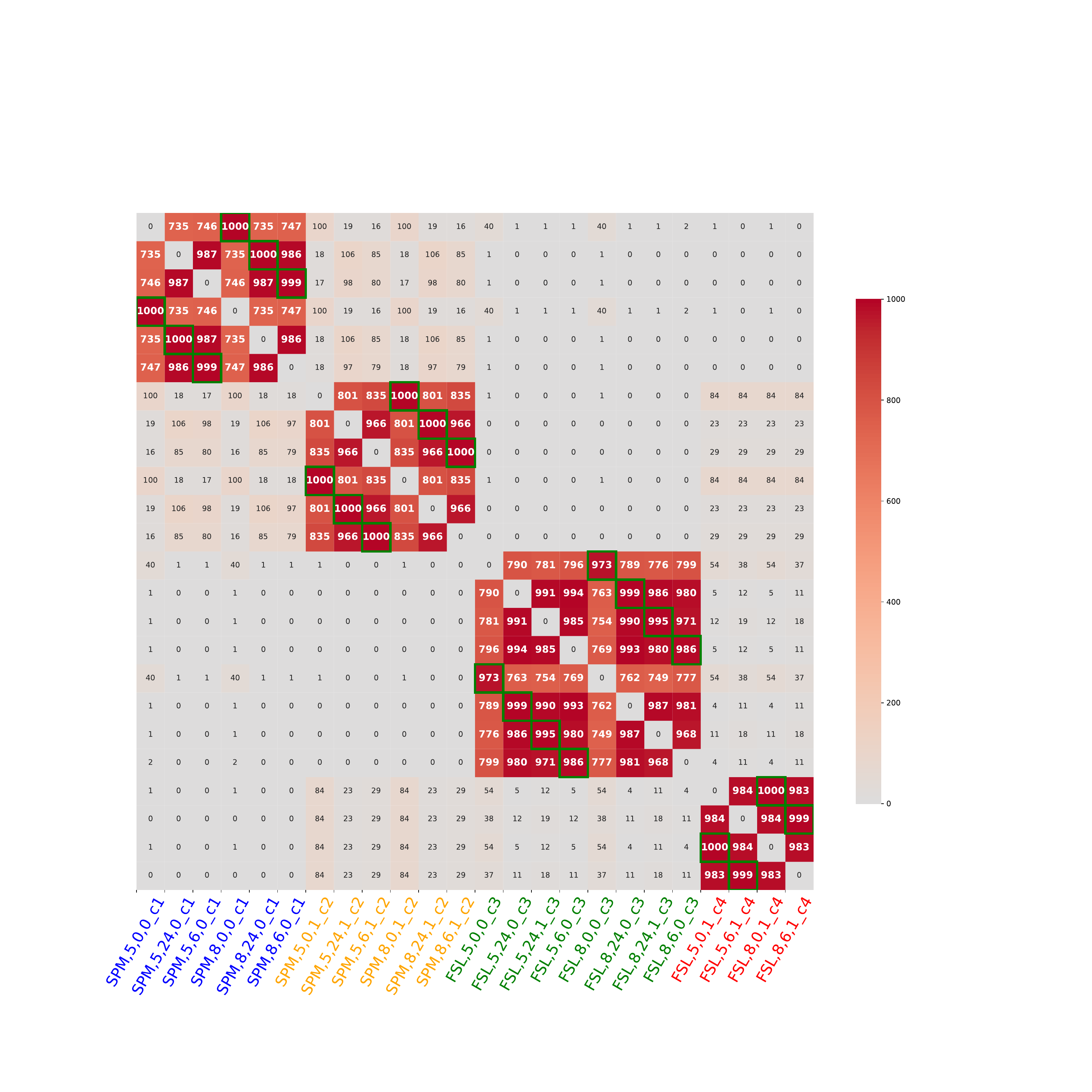}
    \caption{Adjacency matrix representing the number of times each pair pipelines belong to the same community across different group-level statistic maps of the contrast right hand.}
    \label{fig:fig2}
\end{figure}

The adjacency matrix representing the number of times each pair of pipelines belonged to the same community across different group-level statistic maps of the contrast \emph{right-hand} is shown in Fig. \ref{fig:fig2}. The graph corresponding to this adjacency matrix was partitioned using the Louvain community algorithm and 4 communities were identified. These communities correspond to groups of pipelines that are frequently located in the same community across groups of subjects (i.e., that give similar results for a high number of groups). The partitioning of this graph achieves a modularity of $0.64$ (modularity~\cite{Blondel_2008} takes values between $-0.5$ and $1$ and considered high above $0.3$). 

We can see that pipelines inside each partition share specific parameters, these parameters are the main factors that distinguish pipelines between communities, i.e. that drives the variability of the pipeline space. Here, in each community, we can find pipelines with the same software package and the same use of HRF derivatives. This means that for this contrasts, pipelines sharing these parameters give closer results than pipelines with other ones. 

Inside communities, pairs of pipelines show a large number of co-occurrence in the same community across groups (more than 700 for all pairs of pipelines in each community). This means that the relationships observed between pipelines results are stable across different groups of subjects. In particular, pairs of pipelines sharing all parameters except smoothing kernel FWHM are more than 99\% of the time identified in the same community (see green highlight in Fig.~\ref{fig:fig2}). For instance, pipelines `spm-5-0-0' and `spm-8-0-0' are located in community 1 for all groups of subjects. 

\subsection{Communities for contrast \textit{right-hand}}
\begin{figure}[htb]
    \centering
    \includegraphics[width=8.5cm]{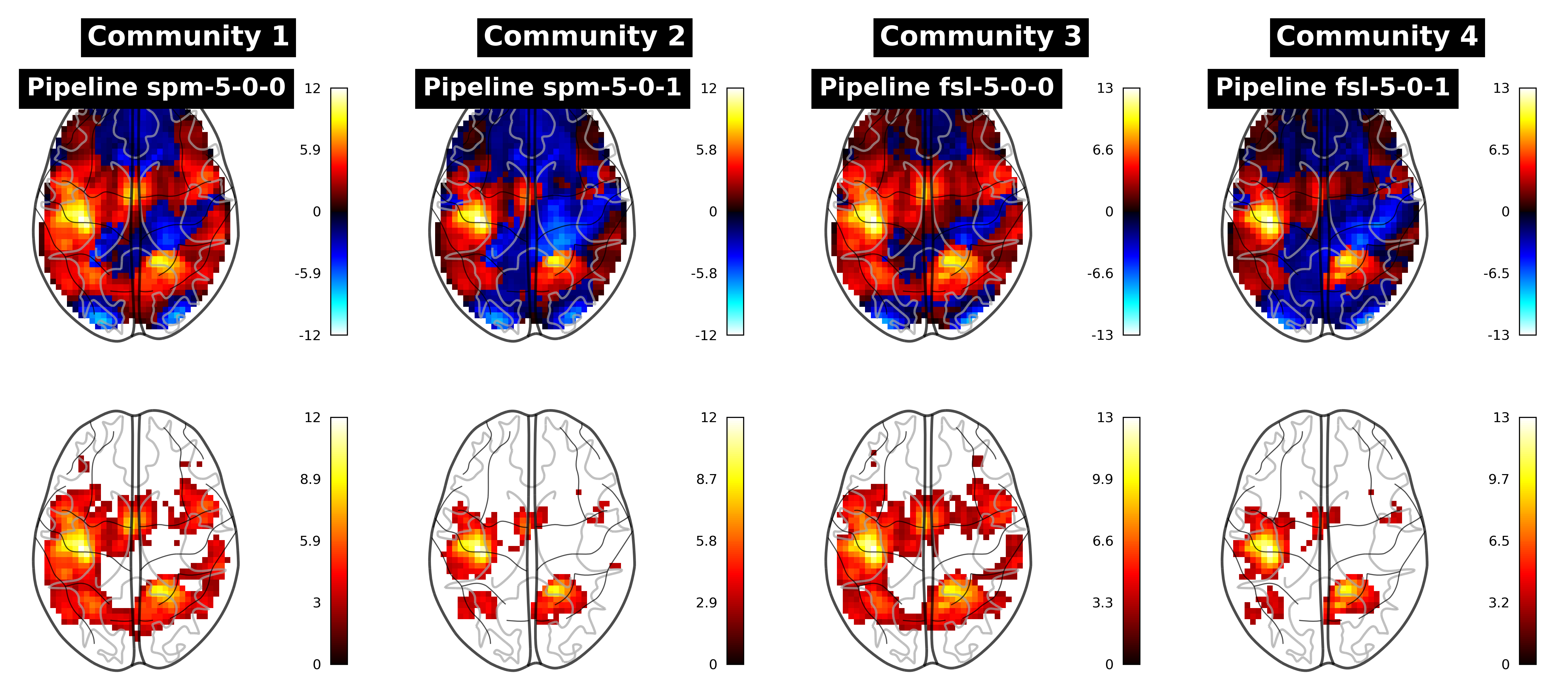}
    \caption{Mean statistic map across groups of subjects for a representative pipeline of each community for the contrast right-hand. Unthresholded maps (upper) and thresholded maps (lower) with FDR-corrected $p < 0.05$.}
    \label{fig:fig4}
\end{figure}

Mean unthresholded (upper) and thresholded (lower) statistic maps of a representative pipeline of the different communities identified for the contrast \textit{right-hand} are displayed in Fig.\ref{fig:fig4}. Mean maps of other pipelines per communities are available in supplementary materials~\cite{supp-data}. This representative pipeline was arbitrarily chosen, other pipelines of each community show similar activation patterns. The global activation patterns are similar across communities, but the activation area is larger for the pipeline of communities 1 and 3. These communities are composed of pipelines that do not include HRF derivatives. We can suppose that this parameter has an impact on the number of significant voxels detected in the analysis. This observation is confirmed by the number of activated voxels in the thresholded maps of the pipelines inside each community  (Tab.\ref{tab:tab1}). Statistic maps of the representative pipeline of communities 1 and 3 show a high number of activated voxels ($N=2,786$ and $2,539$) compared to communities 2 and 4 ($N=796$ and $727$). The numbers of activated voxels inside the ROI of the Primary Motor Cortex are more similar between communities but remain more elevated in communities 1 and 3.

These maps also show that pairs of communities can have similar activation area. We can suppose that the pipelines of these pairs of communities are closer to each other than to the ones from other communities. This suppose that there are distant \& close pipelines (inside vs outside a community), but also distant \& close communities. In this case, pipelines sharing the same use of HRF derivatives (community 1 and 3) seem closer than those having different use of HRF derivatives but the same software package (community 1 and 2).

\begin{table}[h]
    \centering
    \caption{Mean number of activated voxels in the thresholded mean statistic maps of the representative pipeline of each community (1st row) and inside the ROI of the Primary Motor Cortex (2nd row) for the contrast right-hand.}
    \label{tab:tab1}
    \begin{tabular}{|c|c|c|c|c|}
        \hline
        \textbf{Community} &  \textbf{1} & \textbf{2} & \textbf{3} & \textbf{4} \\
        \hline 
        \textbf{Whole maps} & 2,786 & 796 & 2,539 & 727 \\
        \hline
        \textbf{ROI} & 382 & 252 & 337 & 215 \\ 
        \hline
    \end{tabular}
\end{table}

\subsection{Characteristics of the pipeline space for contrast \textit{right-foot}}

\begin{figure}[htb!]
    \centering
    \includegraphics[width=8.5cm]{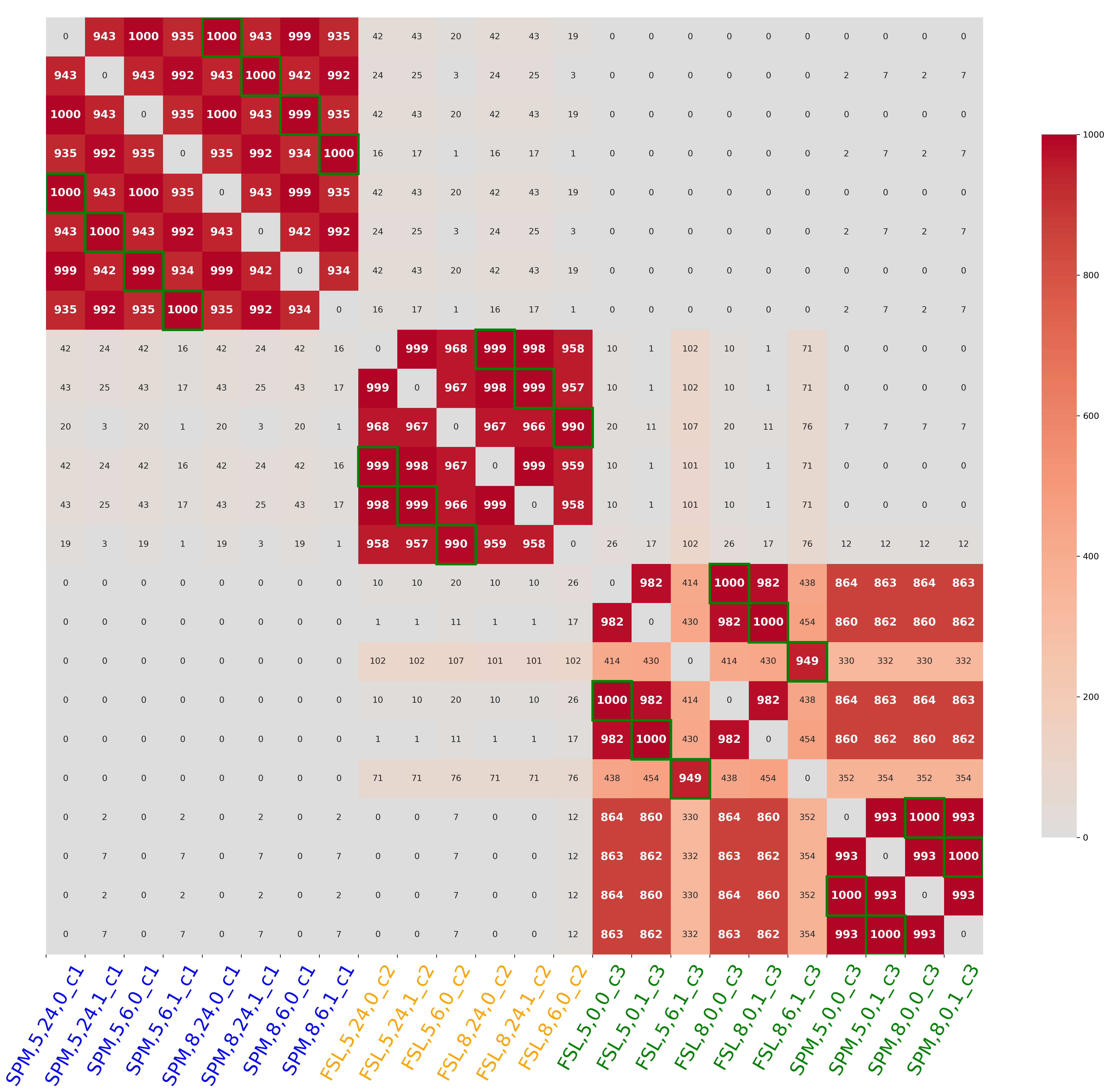}
    \caption{Adjacency matrix representing the number of times each pair pipelines belong to the same community across different group-level statistic maps of the contrast right foot.}
    \label{fig:fig3}
\end{figure}

Fig.~\ref{fig:fig3} shows the adjacency matrix for the contrast \textit{right-foot}. For this contrast, only 3 communities are identified and the repartition of pipelines inside the communities differ compared to the one observed for the contrast \textit{right-hand}. In Fig.~\ref{fig:fig2}, for contrast \textit{right-hand}, communities are composed of pipelines with different software packages (communities 1 vs 3) and different use of HRF derivatives (communities 2 vs 4). For the contrast \textit{right-foot}, the main factors that drives the repartition of pipelines in communities do not seem to be related to the software package: communities 1 and 2 contain pipelines from different software packages, but community 3 is composed of both SPM and FSL pipelines. In this case, the use of different numbers of motion regressors seems to have a larger impact on community identification (pipelines with 6 or 24 motion regressors are located in communities 1 and 2 vs. 0 or 6 motion regressors in community 3). 

This suppose that the relationships between pipeline results vary between contexts: here, cognitive paradigm and groups of subjects. In supplementary materials~\cite{supp-data}, we also show the adjacency matrices obtained for the contrasts \textit{left-hand} and \textit{left-foot}, i.e. same cognitive paradigm as those presented in Fig.~\ref{fig:fig2} and~\ref{fig:fig3} but located in the controlateral brain hemisphere. Communities identified for these left paradigms are similar to those observed for the corresponding right ones. This suppose that pipeline behaviors, and thus relationships between different pipelines, are related to the effect to detect (here, activation of the brain when performing a motor action with the hand or the foot) and the size of this effect. 

We can also observe that the detected communities are slightly less stable across groups of participants for contrast \textit{right-foot}, in particular for community 3 some pairs of pipelines show a number of co-occurence in the same community of less than 500 out of 1,000. 

\begin{figure}[htb]
    \centering
    \includegraphics[width=8.5cm]{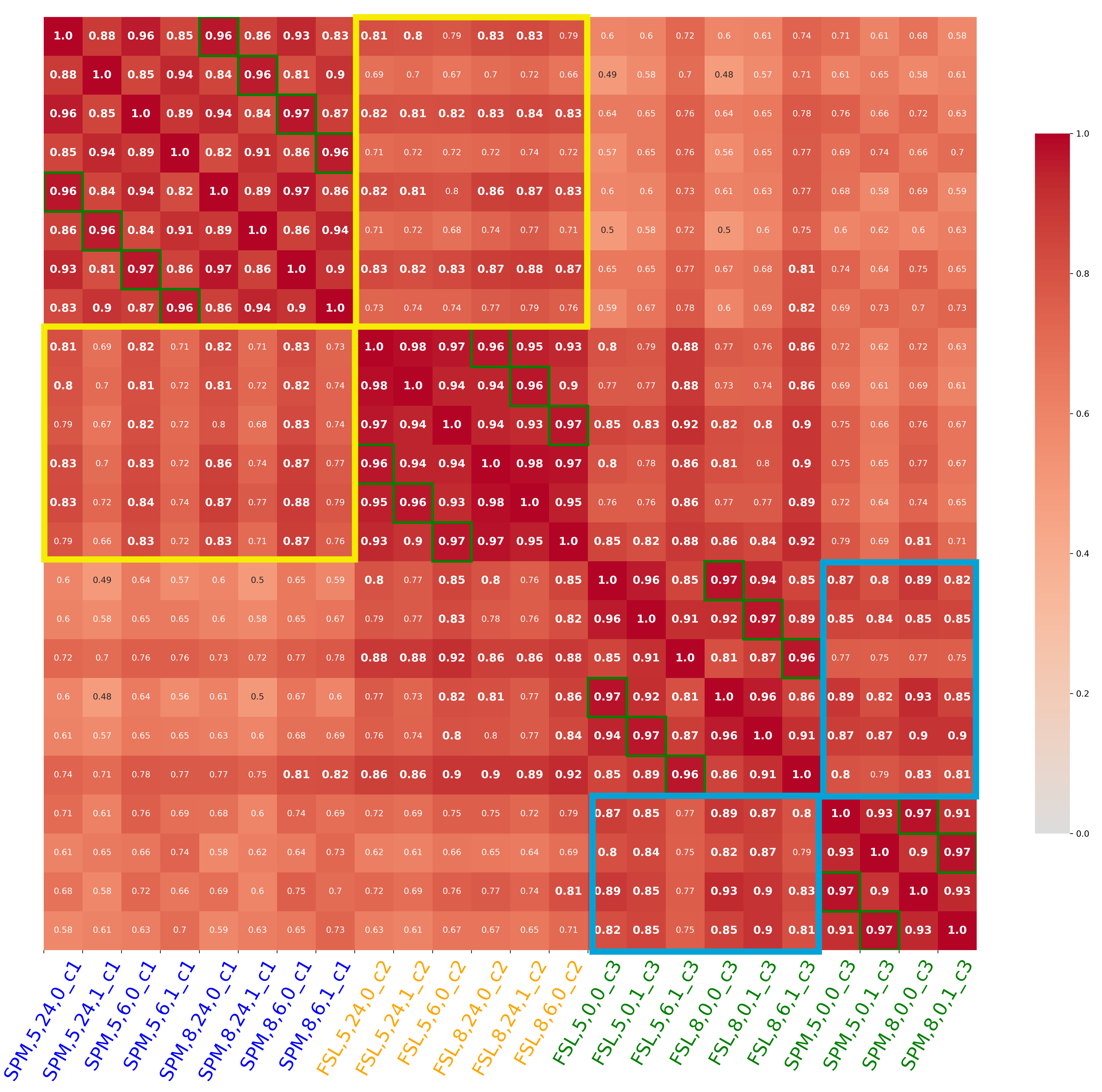}
    \caption{Adjacency matrix representing the mean correlations across groups between statistic maps of each pair pipelines for the contrast right foot. Correlations between statistic maps of pipelines located in community 1 and community 2 are shown in a \textbf{yellow} box. Correlations between statistic maps of pairs pipelines located in community 3 that have a low number of co-occurence in the same community are shown in a \textbf{blue} box.}
    \label{fig:corr-matrix}
\end{figure}

To explore this findings, we looked at the matrix of mean correlations pipeline-to-pipeline across groups (Fig.~\ref{fig:corr-matrix}). Pipelines of community 3 for which the number of co-occurence in the communities with other pipelines is low are highlighted in blue. We can see that correlations between these pipelines are lower than other correlations inside the community, for instance: pipelines `fsl,5,6,1' and `spm,8,0,1' are both located in community 3 for only 55 groups out of 1,000 and the mean correlation between their statistic maps is of 75\%. In comparison, pipelines `spm,8,0,0' and `spm,8,0,1' are co-located in community 3 for 972 groups and the correlation between their maps is of 93\%. These observations might explains the low stability observed in this community. 

This matrix also shows that results of pipelines inside a community can be close to those of a community but distant from those of another. Here, statistic maps of pipelines in community 1 seem closer to the ones of community 2 than to those of community 3. However, this does not impact the stability of relationships since between-communities correlations (around 0.8) are still lower than intra-communities correlations (0.9). 

\subsection{Communities for the contrast \textit{right-foot}}
\begin{figure}[ht!]
    \centering
    \includegraphics[width=7.5cm]{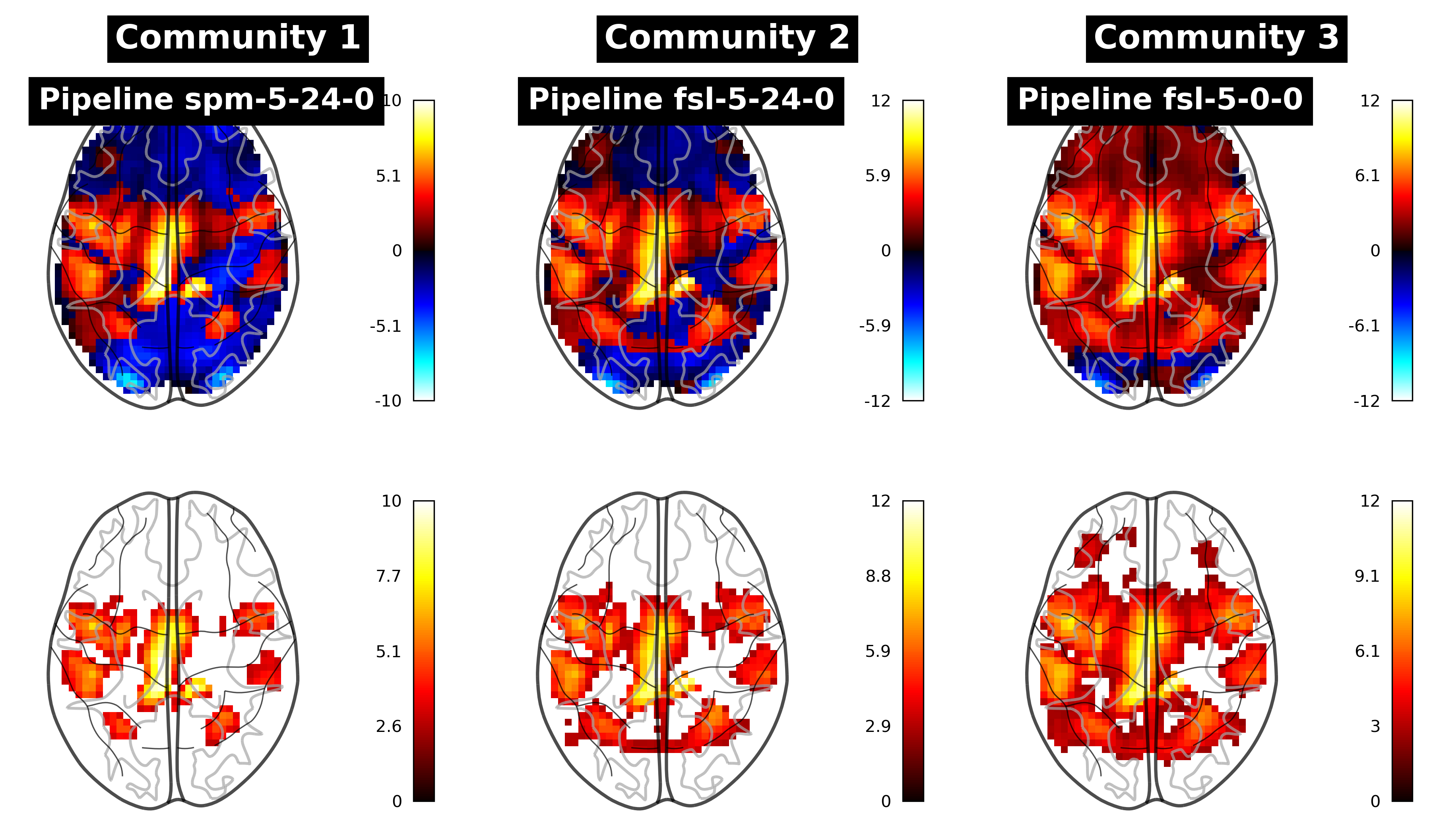}
    \caption{Mean statistic map across groups of subjects for a representative pipeline of each community for the contrast right-foot. Unthresholded maps (upper) and thersholded maps (lower) with FDR-corrected $p < 0.05$.}
    \label{fig:fig5}
\end{figure}

Mean unthresholded (upper) and thresholded (lower) statistic maps of a representative pipeline of each community identified for the contrast \textit{right-foot} are also displayed in Fig.\ref{fig:fig5}. Observations are similar as those made for maps of the contrast \textit{right-hand} (Fig.~\ref{fig:fig4}), but the differences between statistic maps of communities in terms of size of activation is larger than for the right-hand contrast. The size of activation areas seems to be an important criteria to group pipelines in communities and seems to be related to different pipeline parameters depending on the context, here on the cognitive paradigm and the size of the effect to detect. 

\section{Discussion} \label{sec:discussion}
In this work, we used community detection algorithms to explore the relationships between statistic maps obtained with different pipelines in task-fMRI. Our goal was to gain a better understanding of the relationships between the results of different pipelines and of the stability of these relationships in different contexts (i.e. across group of participants and cognitive paradigm). We were able to identify communities of pipelines that were giving close results across different groups of subjects, but not across cognitive paradigm. Pipelines inside each community shared specific parameters values: for instance, same software package and use of HRF derivative for the communities identified in contrast \textit{right-hand} and same use of motion regressors for contrast \textit{right-foot}. Identification of such parameters that drives the relationships between pipeline results, is crucial to select the pipelines to explore in multiverse analyzes and mega-analyses. 

One of the main findings of this work is that the relationships between pipelines is not stable and can depend on the contrast and on the group of subject studied. In particular, for specific pairs of pipelines, the relationship between their results can vary across groups of subjects with a number of co-occurence in the same community less than 500/1,000. This means that two pipelines can give very close results for a group of subjects and more distant ones for another group. Relationships between pipeline results are even more variable when comparing different cognitive paradigm, as two pipelines can be identified in different communities (i.e. giving distant results) for a contrast and located in the same one (i.e. giving close results) for another contrast. Here, the communities of pipelines identified for the contrast \textit{right-hand} were different from those identified for \textit{right-foot}, but these were similar to those identified for the contrast \textit{left-hand}. As we may suppose that controlateral paradigms are similar, this suggests that pipeline behaviors, and thus relationships between pipelines results, might depend on specific characteristics of the paradigm, for instance of the size of the effect to detect, as previous findings in the literature showed a stronger brain activity detected by functional MRI during the execution of finger movements than toe or tongue movements~\cite{ehrsson_imagery_2003}.

This relative instability of the relationships between pipeline results puts into question the ability to learn a mapping between pipelines. Indeed, to facilitate data re-use with maps coming from different pipelines, a possible solution would be to learn a mapping between pairs of pipelines to convert statistic maps. This can be done through style transfer~\cite{isola_image--image_2018}, a widely used framework in the field of computer vision. However, this requires a stable relationship between the data from the two pipelines to train the style transfer model and for inference, for instance if we want to apply a model trained on data from a cognitive paradigm on another one. Our findings suggest that such models might be hard to generalize to different datasets, in particular if the data explore other cognitive paradigms than the ones seen during training.

Pipelines statistic maps in communities shared similar activation patterns. In particular, we found that the main distinguishing factor between communities seemed to be related to the size of the activation area, in particular for communities 1-3 and 2-4 for the contrast \emph{right-hand}. In this context, regarding the composition of communities, we could suppose that the use of HRF derivatives in the pipelines led to a more restricted activation area in the resultant statistic maps. However, statistic maps of representative pipelines of communities with different software packages (communities 1-3 and 2-4) showed very similar activation patterns but differed in terms of the scale of statistic values. This can be explained by differences in terms of method implementation between software packages, e.g. pre-whitening methods which have been shown to impact the number of significant voxels~\cite{olszowy_accurate_2019}. FSL analyses tend to lead to higher statistical values, which might explain the lower correlations between the maps of pipelines coming from different software packages. Thus, we could conclude that these pipelines parameters had an impact on the size of the activation area and on the scale of statistical values, which were sufficiently different in pipelines results to group them into communities. 

By comparing communities based on the correlations between statistic maps of pairs of pipelines, we showed that results of pipelines in communities were more similar to those of some communities than to those of others. This finding could explain the lack of stability observed across groups in some case: if two communities show very similar patterns and knowing that specific properties of the data might influence the behavior of a pipeline, it is likely that one pipeline can be sometimes identified in another community. We suppose that this observation on the lack of stability of data and the similarities of some communities might be related to the pipelines and parameters we analyzed.  

Indeed, the main limitation of our work is the use of a constrained set of pipelines implementation. Indeed, the HCP multi-pipeline dataset contains statistic maps output from 24 different pipelines that varies for 4 parameters (software packages, smoothing kernel FWHM, number of motion regressors, and use of HRF derivatives). These pipelines were chosen to represent typical pipelines found in the literature~\cite{carp_plurality_2012}. We also selected these pipelines to represent parameters that have been shown to impact the results when using a different value~\cite{cignetti_pros_2016, carp_plurality_2012}. As we wanted to explore the stability of the pipeline space across different contexts, the 1,000 groups and the 5 contrasts of the motor task present in this dataset were a major advantage. In future work, it would be interesting to explore other datasets, such as the statistic maps resulting from the NARPS many-analyst study~\cite{botvinik2020variability} (one group-level statistic maps for 70 pipelines and 9 research hypotheses), which would allow us to explore a less constrained set of pipelines and a different cognitive paradigm for which the effect size is presumably lower (mixed gamble task). 

\section{Conclusion}

In this study, we explored the fMRI analytical space and its stability across different contexts (contrast, groups of participants). We found that distance between pipelines were mostly driven by the size of the activation areas, which depends on specific characteristics of the paradigm and by the interaction of the effect to detect in this paradigm with some specific pipeline parameters, leading to an instability of the pipeline space. In future work, this workflow may be used with other sets of pipelines and other paradigms to assess the generalizability of our results. These results could thus be used to tackle analytical variability in fMRI analyses with, for instance, the selection of representative pipelines or of pipelines having stable relationships. 

\section*{Acknowledgment}

This work was partially funded by Region Bretagne (ARED MAPIS) and Agence Nationale pour la Recherche for the program of doctoral contracts in artificial intelligence (project ANR-20-THIA-0018). \\
Data were provided by the Human Connectome Project, WU-Minn Consortium (Principal Investigators: David Van Essen and Kamil Ugurbil; 1U54MH091657) funded by the 16 NIH Institutes and Centers that support the NIH Blueprint for Neuroscience Research; and by the McDonnell Center for Systems Neuroscience at Washington University.\\
We thank Jeremy Lefort-Besnard for its advice and its code on Louvain communities.

\bibliographystyle{IEEEbib}
\bibliography{refs}

\begin{thebibliography}{10}

\bibitem{firat_deep_2014}
Orhan Firat, Like Oztekin, and Fatos T.~Yarman Vural,
\newblock ``Deep learning for brain decoding,''
\newblock in {\em 2014 {IEEE} International Conference on Image Processing
  ({ICIP})}. 2014, pp. 2784--2788, {IEEE}.

\bibitem{yin_deep_2022}
Wutao Yin, Longhai Li, and Fang-Xiang Wu,
\newblock ``Deep learning for brain disorder diagnosis based on {fMRI}
  images,''
\newblock {\em Neurocomputing}, vol. 469, pp. 332--345, 2022.

\bibitem{carp_secret_2012}
Joshua Carp,
\newblock ``The secret lives of experiments: {Methods} reporting in the {fMRI}
  literature,''
\newblock {\em NeuroImage}, vol. 63, no. 1, pp. 289--300, 2012.

\bibitem{botvinik2020variability}
Rotem Botvinik-Nezer, Felix Holzmeister, Colin~F Camerer, Anna Dreber, Juergen
  Huber, Magnus Johannesson, Michael Kirchler, Roni Iwanir, Jeanette~A Mumford,
  R~Alison Adcock, et~al.,
\newblock ``Variability in the analysis of a single neuroimaging dataset by
  many teams,''
\newblock {\em Nature}, vol. 582, no. 7810, pp. 84--88, 2020.

\bibitem{bowring_maumet}
Alexander Bowring, Camille Maumet, and Thomas~E. Nichols,
\newblock ``Exploring the impact of analysis software on task fmri results,''
\newblock {\em Human Brain Mapping}, vol. 40, no. 11, pp. 3362--3384, 2019.

\bibitem{dafflon2022guided}
Jessica Dafflon, Pedro F.~Da~Costa, Franti{\v{s}}ek V{\'a}{\v{s}}a, Ricardo~Pio
  Monti, Danilo Bzdok, Peter~J Hellyer, Federico Turkheimer, Jonathan
  Smallwood, Emily Jones, and Robert Leech,
\newblock ``A guided multiverse study of neuroimaging analyses,''
\newblock {\em Nature Communications}, vol. 13, no. 1, pp. 3758, 2022.

\bibitem{ioannidis_why_2008}
John P.~A. Ioannidis,
\newblock ``Why {Most} {Discovered} {True} {Associations} {Are} {Inflated}:,''
\newblock {\em Epidemiology}, vol. 19, no. 5, pp. 640--648, 2008.

\bibitem{simmons_false-positive_2011}
Joseph~P. Simmons, Leif~D. Nelson, and Uri Simonsohn,
\newblock ``False-positive psychology: Undisclosed flexibility in data
  collection and analysis allows presenting anything as significant,''
\newblock {\em Psychological Science}, vol. 22, no. 11, pp. 1359--1366, 2011.

\bibitem{gelman_garden_nodate}
Andrew Gelman and Eric Loken,
\newblock ``The garden of forking paths: {Why} multiple comparisons can be a
  problem, even when there is no “ﬁshing expedition” or “p-hacking”
  and the research hypothesis was posited ahead of time,''
\newblock {\em Department of Statistics, Columbia University}, 2013.

\bibitem{steegen_increasing_2016}
Sara Steegen, Francis Tuerlinckx, Andrew Gelman, and Wolf Vanpaemel,
\newblock ``Increasing {Transparency} {Through} a {Multiverse} {Analysis},''
\newblock {\em Perspectives on Psychological Science: A Journal of the
  Association for Psychological Science}, vol. 11, no. 5, pp. 702--712, 2016.

\bibitem{rolland_towards_2022}
Xavier Rolland, Pierre Maurel, and Camille Maumet,
\newblock ``Towards efficient fmri data re-use: can we run between-group
  analyses with datasets processed differently with spm ?,''
\newblock in {\em {ISBI} 2022 - {IEEE} {International} {Symposium} on
  {Biomedical} {Imaging}}. 2022, pp. 1--4, IEEE.

\bibitem{gorgolewski_neurovaultorg_2015}
Krzysztof~J. Gorgolewski, Gael Varoquaux, Gabriel Rivera, Yannick Schwarz,
  Satrajit~S. Ghosh, Camille Maumet, Vanessa~V. Sochat, Thomas~E. Nichols,
  Russell~A. Poldrack, Jean-Baptiste Poline, Tal Yarkoni, and Daniel~S.
  Margulies,
\newblock ``{NeuroVault}.org: a web-based repository for collecting and sharing
  unthresholded statistical maps of the human brain,''
\newblock {\em Frontiers in Neuroinformatics}, 2015.

\bibitem{germani:inserm-04531405}
Elodie Germani, Camille Maumet, and Elisa Fromont,
\newblock ``{Mitigating analytical variability in fMRI results with style
  transfer},''
\newblock preprint: https://inserm.hal.science/inserm-04531405, 2024.

\bibitem{parsons_subspace_2004}
Lance Parsons, Ehtesham Haque, and Huan Liu,
\newblock ``Subspace clustering for high dimensional data: a review,''
\newblock {\em ACM SIGKDD Explorations Newsletter}, vol. 6, no. 1, pp. 90--105,
  2004.

\bibitem{hcp_multi_pipeline}
Elodie Germani, Elisa Fromont, Pierre Maurel, and Camille Maumet,
\newblock ``{The HCP multi-pipeline dataset: an opportunity to investigate
  analytical variability in fMRI data analysis},''
\newblock preprint: https://inserm.hal.science/inserm-04356768, Dec. 2023.

\bibitem{hcp_overview}
D.C. {Van Essen}, K.~Ugurbil, E.~Auerbach, D.~Barch, T.E.J. Behrens,
  R.~Bucholz, A.~Chang, L.~Chen, M.~Corbetta, S.W. Curtiss, S.~{Della Penna},
  D.~Feinberg, M.F. Glasser, N.~Harel, A.C. Heath, L.~Larson-Prior, D.~Marcus,
  G.~Michalareas, S.~Moeller, R.~Oostenveld, S.E. Petersen, F.~Prior, B.L.
  Schlaggar, S.M. Smith, A.Z. Snyder, J.~Xu, and E.~Yacoub,
\newblock ``The human connectome project: A data acquisition perspective,''
\newblock {\em NeuroImage}, vol. 62, pp. 2222--2231, 2012.

\bibitem{penny2011statistical}
William~D Penny, Karl~J Friston, John~T Ashburner, Stefan~J Kiebel, and
  Thomas~E Nichols,
\newblock {\em Statistical parametric mapping: the analysis of functional brain
  images},
\newblock Elsevier, 2011.

\bibitem{jenkinson2012fsl}
Mark Jenkinson, Christian~F Beckmann, Timothy~EJ Behrens, Mark~W Woolrich, and
  Stephen~M Smith,
\newblock ``Fsl,''
\newblock {\em Neuroimage}, vol. 62, no. 2, pp. 782--790, 2012.

\bibitem{Blondel_2008}
Vincent~D Blondel, Jean-Loup Guillaume, Renaud Lambiotte, and Etienne Lefebvre,
\newblock ``Fast unfolding of communities in large networks,''
\newblock {\em Journal of Statistical Mechanics: Theory and Experiment}, vol.
  2008, no. 10, pp. P10008, oct 2008.

\bibitem{code_swh}
Elodie Germani, Elisa Fromont, and Camille Maumet,
\newblock ``Software heritage archive for the gitlab repository
  "pipeline\_distance".,''
  \url{https://archive.softwareheritage.org/browse/origin/directory/?origin_url=https://gitlab.inria.fr/egermani/pipeline_distance},
  2023.

\bibitem{abraham_machine_2014}
Alexandre Abraham, Fabian Pedregosa, Michael Eickenberg, Philippe Gervais,
  Andreas Mueller, Jean Kossaifi, Alexandre Gramfort, Bertrand Thirion, and
  Gael Varoquaux,
\newblock ``Machine learning for neuroimaging with scikit-learn,''
\newblock {\em Frontiers in Neuroinformatics}, 2014.

\bibitem{supp-data}
Elodie Germani, Elisa Fromont, and Camille Maumet,
\newblock ``Supplementary materials for "uncovering communities of pipelines in
  the task-fmri analytical space",''
  \url{https://sigport.org/documents/supplementary-materials-uncovering-communities-pipelines-task-fmri-analytical-space},
  2024.

\bibitem{ehrsson_imagery_2003}
H.~Henrik Ehrsson, Stefan Geyer, and Eiichi Naito,
\newblock ``Imagery of {Voluntary} {Movement} of {Fingers}, {Toes}, and
  {Tongue} {Activates} {Corresponding} {Body}-{Part}-{Specific} {Motor}
  {Representations},''
\newblock {\em Journal of Neurophysiology}, vol. 90, no. 5, pp. 3304--3316,
  2003.

\bibitem{isola_image--image_2018}
Phillip Isola, Jun-Yan Zhu, Tinghui Zhou, and Alexei~A. Efros,
\newblock ``Image-to-{Image} {Translation} with {Conditional} {Adversarial}
  {Networks},'' 2018.

\bibitem{olszowy_accurate_2019}
Wiktor Olszowy, John Aston, Catarina Rua, and Guy~B. Williams,
\newblock ``Accurate autocorrelation modeling substantially improves {fMRI}
  reliability,''
\newblock {\em Nature Communications}, vol. 10, pp. 1220, 2019.

\bibitem{carp_plurality_2012}
Joshua Carp,
\newblock ``On the plurality of (methodological) worlds: estimating the
  analytic flexibility of {fMRI} experiments,''
\newblock {\em Frontiers in Neuroscience}, 2012.

\bibitem{cignetti_pros_2016}
Fabien Cignetti, Emilie Salvia, Jean-Luc Anton, Marie-Hélène Grosbras, and
  Christine Assaiante,
\newblock ``Pros and {Cons} of {Using} the {Informed} {Basis} {Set} to
  {Account} for {Hemodynamic} {Response} {Variability} with {Developmental}
  {Data},''
\newblock {\em Frontiers in Neuroscience}, vol. 10, 2016.

\end{thebibliography}

\end{document}